\documentclass{llncs}
\usepackage{graphicx}
\usepackage{mathtools}
\usepackage{amsmath}

\begin{document}

\title{siftservice.com - Turning a Computer Vision algorithm into a World Wide Web Service}
\titlerunning{Hamiltonian Mechanics}  
%
\author{Ahmad Pahlavan Tafti\inst{1} \and Hamid Hassannia\inst{2}
 \and Zeyun Yu\inst{1}}
\authorrunning{Ivar Ekeland et al.} 
%

%
\institute{Department of Computer Science, University of Wisconsin -Milwaukee, WI, USA
\and
IEEE Member, Uppsala, Sweden}

\maketitle              

\begin{abstract}
Image features detection and description is a longstanding topic in computer vision and pattern recognition areas. The Scale Invariant Feature Transform (SIFT) is probably the most popular and widely demanded feature descriptor which facilitates a variety of computer vision applications such as image registration, object tracking, image forgery detection, and 3D surface reconstruction. This work introduces a Software as a Service (SaaS) based implementation of the SIFT algorithm which is freely available at {\bf http://siftservice.com} for any academic, educational and research purposes. The service provides application-to-application interaction and aims Rapid Application Development (RAD) and also fast prototyping for computer vision students and researchers all around the world. An Internet connection is all they need!

\end{abstract}
\section{Introduction}
The Scale Invariant Feature Transform (SIFT) \cite{sift1} has been the topic of many computer vision projects due to its ability to detect feature points which are invariant to image rotation, translation, and scaling. This is obviously evident from Figure 1 which presents the journal and conference papers related to the SIFT algorithm which have been published in Elsevier from 2009 to 2014. 
While several scientists have entered the development of SIFT based algorithm \cite {sift3}, \cite {sift4}, \cite {sift5}, \cite {sift6}, providing a service-based SIFT algorithm over the Internet protocols has not yet been developed. Having a Software as a Service (SaaS) based \cite{saas1}, \cite{saas4}  SIFT algorithm can be fulfilled the following objectives:

\begin{figure}
  \caption{Number of Elsevier publications over last six years. Results achieved by submitting the query "SIFT algorithm" from Elsevier website at http://www.sciencedirect.com.}
  \centering
  \includegraphics [height=5cm, width=9cm]{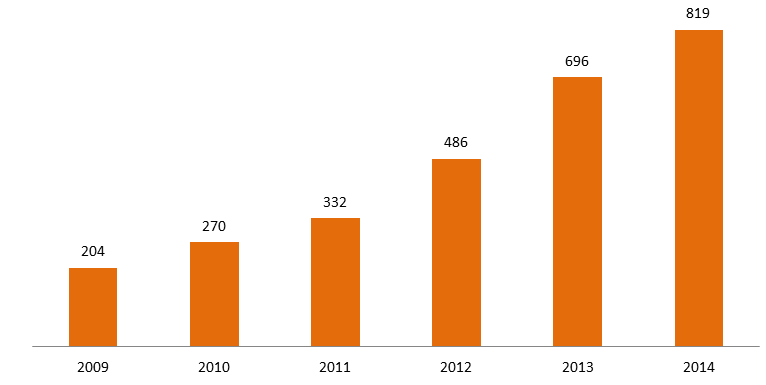}
\end{figure}

\begin{itemize} \itemsep.2em

\item [$\bullet$] To facilitate Rapid Application Development (RAD) and fast prototyping for computer vision students and scientists by a tiny service available on the Internet.

\item [$\bullet$] To provide application-to-application interaction for a highly demanded computer vision algorithm.

\item [$\bullet$] To make the SIFT algorithm available through both human-oriented and 	application-oriented interfaces.

\end{itemize}

In this contribution, we novel design and develop a SaaS based architecture to implement a platform independent and reusable software component for the SIFT algorithm. The service makes a new way to produce and exchange of SIFT information over the standard Internet protocols. Based on its flexibility and availability, it brings an opportunity for computer vision researchers who want to apply the SIFT algorithm on their own images without the need of any programming and installing any software applications. An Internet connection is all we need! 

This work initiates a study of SaaS based architecture for an essential computer vision technique. We expect the work can bridge the gap between computer vision applications and World Wide Web services, stimulating more interests from the computer vision community to the SaaS architecture and the Internet of Things (IoT) areas. Our contribution could be considered as a preliminary step towards the {\bf Computer Vision as a Service}. 

The rest of the paper is arranged as follows. We first give a brief explanation of the system and service design in Section 2. The experimental results along with an application of the SIFT algorithm and the impact of the service are shown in Section 3. Conclusion is presented in Section 4.

\section{System and Service Design}

SIFT algorithm has basically four main stages, namely: 1) Scale-space construction using Difference-of-Gaussian (DoG), 2) Stable features localization, 3) Gradient orientation computation and magnitude assignment, and 4) Feature descriptors extraction \cite{sift1}. 
The first stage is to construct a DoG image pyramid to determine the potential feature points in an image. As it is shown in equation (1), to create Gaussian filtered image, we should convolve the input image $\mathit I(x,y)$ with a Gaussian kernel $\mathit K(x,y; \sigma)$ (equation (2)), where $\mathit \sigma$ is the scale of the Gaussian kernel, and $\mathit conv2$ is 2D convolution operation. Various Gaussian blurred images in different scales are produced and DoGs are computed from neighbors in the scale space. Then candidate feature points would be  detected by discovering exterma in the DoG images which are locally minima or maxima in scale and space.

\begin{equation}
   G (x,y; \sigma)  = conv2(I(x,y),K(x,y,\sigma))
\end{equation}
\begin{equation}
K (x,y; \sigma) = \frac{1}{ {2\pi }{\sigma^2 }}e^{{{ - \left( {x^2 + y^2} \right) } \mathord{\left/ {\vphantom {{ - \left( {x - \mu } \right)^2 } {2\sigma ^2 }}} \right. \kern-\nulldelimiterspace} {2\sigma ^2 }}}
\end{equation}

The next step is to compute the image gradient magnitude and principal orientation to extract associate SIFT descriptor for the a detected feature point. Gradient magnitude as $\mathit m(x,y)$  and principal orientation as $\mathit \theta(x,y)$ are calculated using the following two equations:

\begin{equation}
   \theta (x,y)  = \tan^{-1} ((G(x, y+1)-G(x, y-1))/(G(x+1, y)-G(x-1, y)))
\end{equation}
\begin{equation}
   m (x,y)  = \sqrt {(G(x, y+1)-G(x, y-1))^2+(G(x+1, y)-G(x-1, y))^2}
\end{equation}

Finally, for each feature point, a set of orientation histograms will be created on $ 4 \times 4$ pixel neighborhoods with eight bins each. The size of the descriptor vector can be varied.
The high level service architecture of the siftservice.com is shown in Figure 2.

\vspace{1em}
\begin{figure}
  \caption{siftservice.com: The service architecture.}
  \centering
  \includegraphics [height=8cm, width=10cm] {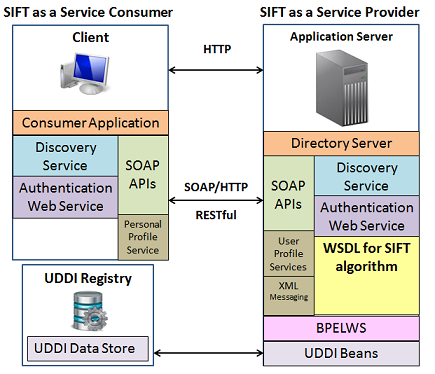}
\end{figure}
\vspace{1em}

The service compromises two disparate machines: a service consumer (Client) and the service provider (Application Server) \cite{saas5}. Main components of each section are presented within the subsystem blocks (Figure 2). 
WSDL stands for Web Service Description Language, and it is an XML-based language for describing SIFT web service and how to access it over the Internet protocols. Discovery Services permit the discovery of the SIFT service. BPELWS stands for Business Process Execution Language for Web Service, and it aims to support the major behaviors of both executable and abstract implementation of the service. 
SOAP stands for Simple Object Access Protocol \cite{saas2}, \cite{saas3}. SOAP is an XML-based Internet protocol which supports exchanging structured information between computers and applications. UDDI which stands for Universal Description, Discovery and Integration is a specification for a distributed registry of the SIFT service \cite{saas1}, \cite{saas2}, \cite{saas3}, \cite{saas4}. An snapshot of siftservice.com is shown in Figure 3.

\section {Experimental Results}

In order to examine the general performance of the system, two experiments were carried out on real digital images. Section 3.1 shows the experimental setup. In Section  3.2, we compare the feature points detection accuracy of the system with the original executable SIFT software implemented by David Lowe \cite{sift2}. In Section 3.3, we analyze and compare matching accuracy between SIFT keypoints in two images using the SIFT as a Service and the original executable SIFT software \cite{sift2}. We then introduce an application of the SIFT algorithm in Section 3.4. The impact of the siftservice.com is presented in Section 3.5.

\subsection {Experimental Setup}
All modules, classes, components, and communication packages of siftservice.com were implemented by Java SE 7. In the server side (Application Server), we used a 64-bit Linux CentOS operating system on a virtual server with 2v cores processor, 100 Mbits/s bandwidth, and 2GB of RAM. In the client side (Client), we employed 64-bit MS Windows 8 operating system with 3.00 GHz Intel Dual core CPU, 2MB cache, and 4GB of RAM. 

\vspace{1em}
\begin{figure}
  \caption{siftservice.com: An snapshot.}
  \centering
  \includegraphics [height=9cm, width=10.5cm]{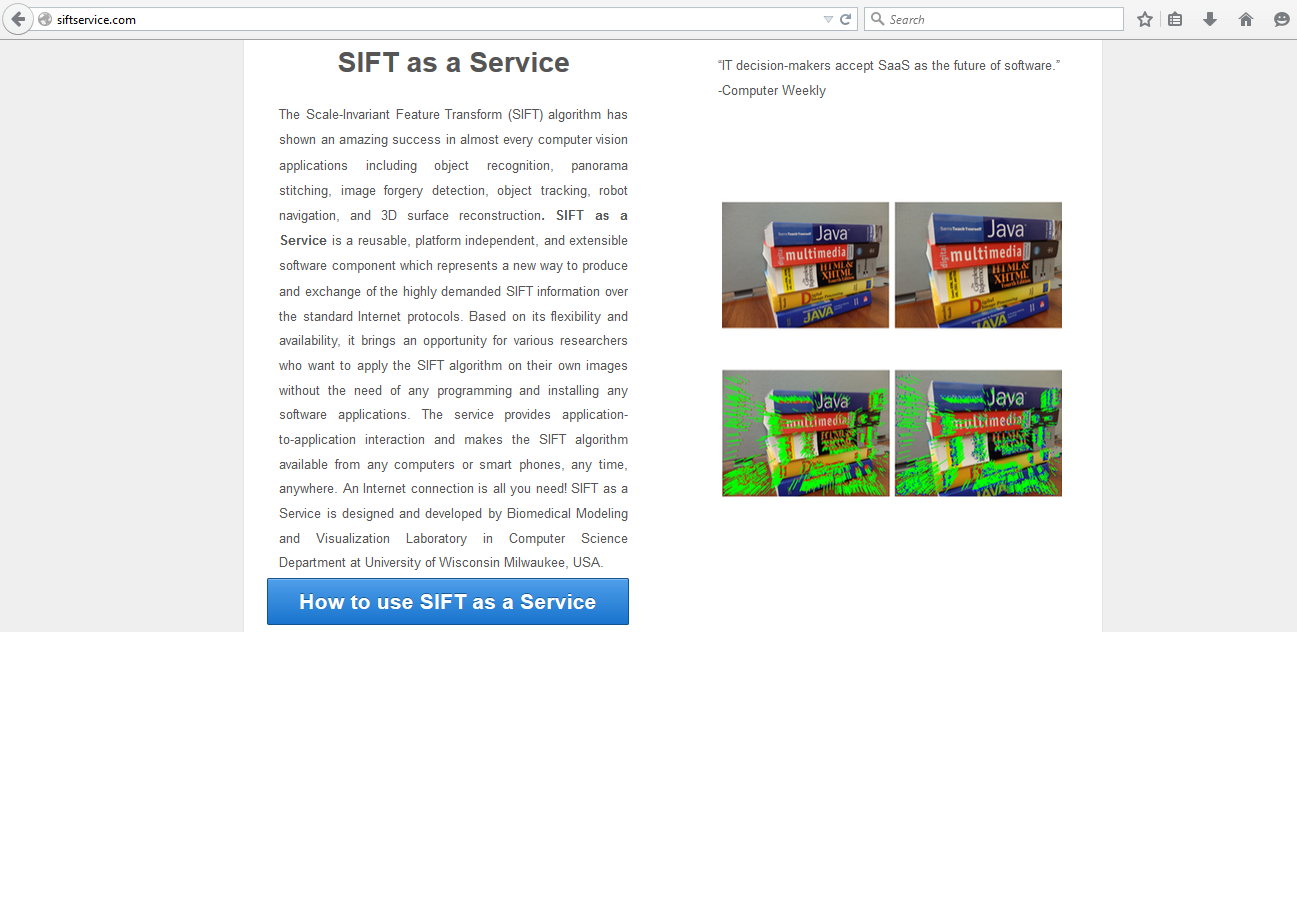}
\end{figure}
\vspace{1em}

\subsection{Accuracy in Feature Points Detection}

In this section we bring up a validation summary on the accuracy and reliability of the proposed system for feature points detection. Figure 4 shows the accuracy in feature points detection. The service produces feature points which are very similar to the original one implemented by Lowe. The error threshold of the SIFT as a Service is less than 1\% in feature points detection.

\vspace{1em}
\begin{figure}
  \caption{Accuracy in Feature Points Detection. (A) Results obtained by the original executable SIFT software implemented by David Lowe (number of feature: 1638). (B) Results obtained by the SIFT as a Service (number of feature: 1627). Image Set: EMS Building.
}
  \centering
  \includegraphics [height=4.0cm, width=10cm]{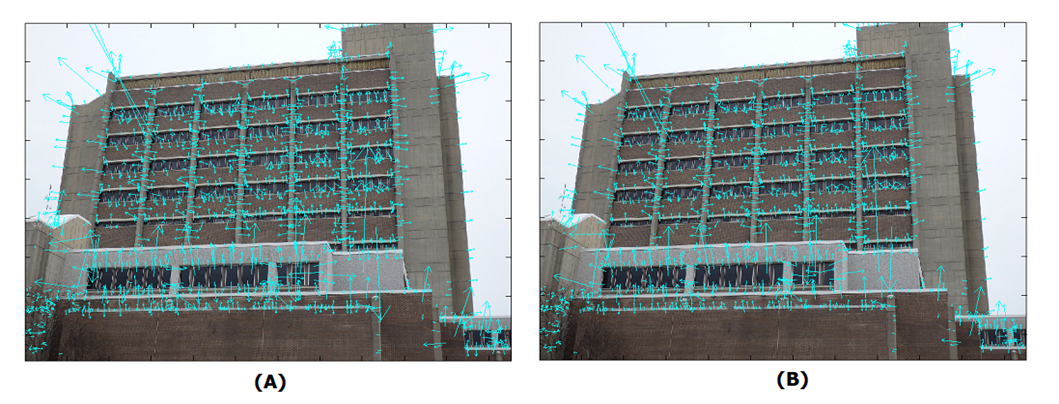}
\end{figure}
\vspace{1em}

\subsection{Accuracy in Points Matching}
Figure 5 shows the points matching accuracy for images "hall01" and "hall02" form the "Hall" images set. We can see that the error threshold for point matching is less than 2\% comparing with the David Lowe's implementation.

\vspace{1em}
\begin{figure}
  \caption{Accuracy in points matching. (A) Results obtained by the original executable SIFT software implemented by David Lowe (number of matches: 166). (B) Results obtained by the SIFT as a Service (number of matches:158). Image Set: Hall.}
  \centering
  \includegraphics [height=4.0cm, width=10cm]{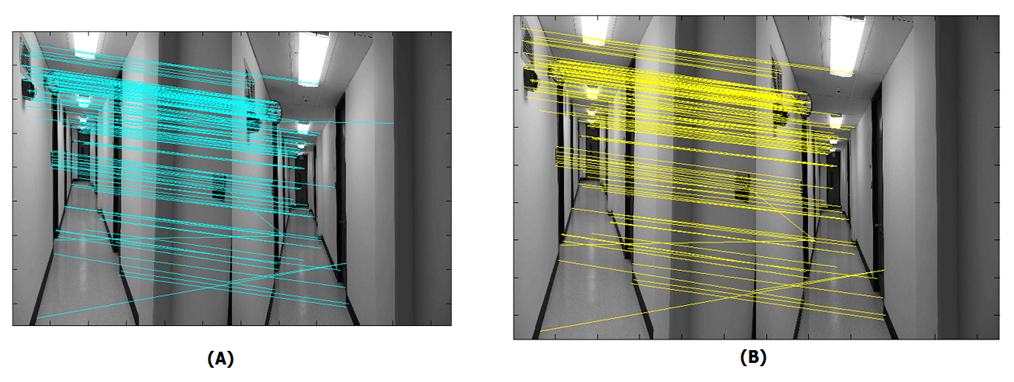}
\end{figure}
\vspace{1em}

\subsection {An Application: 3D Microscopy Vision}
As we mentioned in the abstract, The SIFT algorithm has made in big advance in many areas of computer vision application. An example is 3D surface reconstruction, and in particular 3D microscopy vision. The general pipeline of 3D microscopy vision is presented in Figure 6. The SIFT algorithm is used in step 2 of the pipeline. For further details and information on the proposed method, please refer to \cite{sem1}. We just recall that in that work \cite{sem1}, we employed SURF \cite{surf1} instead of SIFT, but other parts would be same.

\vspace{1em}
\begin{figure}
  \caption{The SIFT algorithm: An application.}
  \centering
  \includegraphics [height=6.5cm, width=10cm]{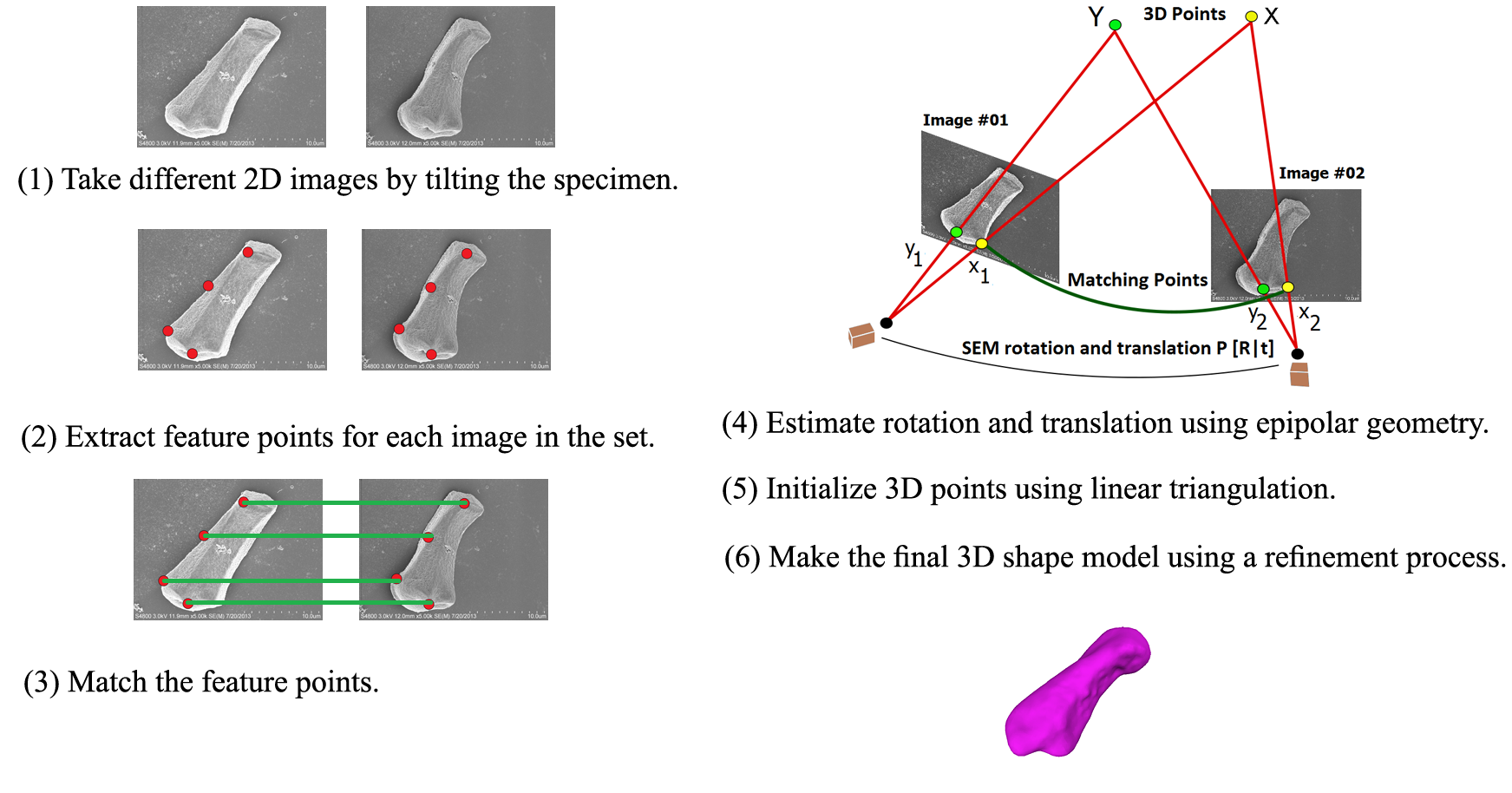}
\end{figure}
\vspace{1em}

\subsection{siftservice.com Imapct}
siftservice.com has been introduced and it is going to be popular for computer vision research community. Figure 7 present the impact of siftservice.com from December 2014 to March 2015.

\vspace{1em}
\begin{figure}
  \caption{siftservice.com impact. Number of users from December 2014 to March 2015.
}
  \centering
  \includegraphics [height=6.0cm, width=10cm]{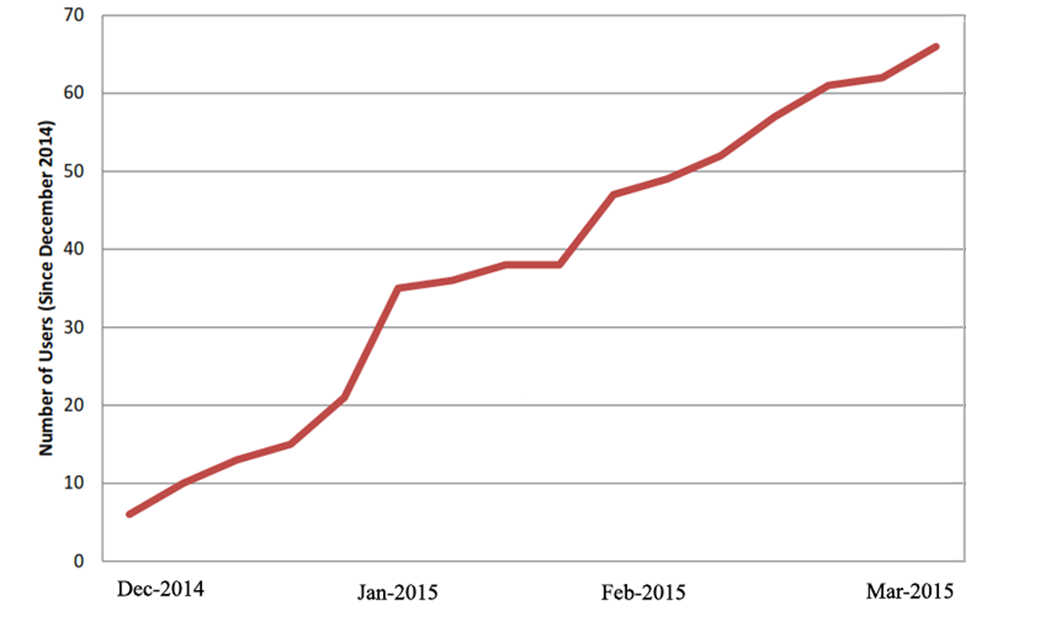}
\end{figure}
\vspace{1em}

 \section{Conclusion}

The SIFT algorithm has been widely used in many computer vision applications. Having SIFT as a Service would provide application-to-application interaction and makes SIFT easier for computer vision researchers and students to Rapid Application Development (RAD), communicating each other applications by a tiny service available in the Internet.  In this project, we have designed and implemented the entire components of the system and addressed some experimental results which show that the service offers promising results. The present work is expected to stimulate more interest and draw attentions from the computer vision community to the fast-growing Software as a Service and the Internet of Things (IoT) area. Our contribution could be considered as a preliminary step towards the {\bf Computer Vision as a Service}. 
The service is freely available at {\bf http://siftservice.com} for any academic, educational and research purposes.

%
%

\end{document}